\documentclass[10pt,twocolumn,letterpaper]{article}
\pdfoutput = 1
\usepackage{cvpr}
\usepackage{times}
\usepackage{epsfig}
\usepackage{graphicx}
\usepackage{amsmath}
\usepackage{amssymb}
\usepackage{subcaption}
\usepackage{tabularx}
\usepackage{multirow,bigdelim}
\usepackage{booktabs}
\usepackage{dblfloatfix}

\usepackage[pagebackref=true,breaklinks=true,letterpaper=true,colorlinks,bookmarks=false]{hyperref}

 \cvprfinalcopy 

\def\cvprPaperID{3406} 

\ifcvprfinal\fi

\begin{document}

\newcommand*\samethanks[1][\value{footnote}]{\footnotemark[#1]}
\title{FA-RPN: Floating Region Proposals for Face Detection} 

\author{\ Mahyar Najibi  $^*$ 
\hspace{3ex}
 Bharat Singh \thanks{Equal Contribution}  \hspace{3ex} Larry S. Davis \\
{\tt\small \{najibi,bharat,lsd\}@cs.umd.edu}}

\maketitle

\begin{abstract}
We propose a novel approach for generating region proposals for performing face-detection. Instead of classifying anchor boxes using features from a pixel in the convolutional feature map, we adopt a pooling-based approach for generating region proposals. However, pooling hundreds of thousands of anchors which are evaluated for generating proposals becomes a computational bottleneck during inference. To this end, an efficient anchor placement strategy for reducing the number of anchor-boxes is proposed. We then show that proposals generated by our network (Floating Anchor Region Proposal Network, FA-RPN) are better than RPN for generating region proposals for face detection. We discuss several beneficial features of FA-RPN proposals like iterative refinement, placement of fractional anchors and changing anchors which can be enabled without making any changes to the trained model. Our face detector based on FA-RPN obtains 89.4\% mAP with a ResNet-50 backbone on the WIDER dataset.
\end{abstract}

\section{Introduction}
Face detection is an important computer vision problem and has multiple applications in surveillance, tracking, consumer facing devices like iPhones \etc. Hence, various approaches have been proposed towards solving it \cite{yang2015facial,zhang2017s,hu2017finding,zhu2017cms,li2016face,sun2017face,zhu2018seeing,ranjan2017hyperface,najibi2017ssh} and successful solutions have also been deployed in practice. So, expectations from face detection algorithms are much higher and error rates today are quite low. Algorithms need to detect faces which are as small as 5 pixels to 500 pixels in size. As localization is essential for detection, evaluating every small region of the image is important. Face detection datasets can have up to a thousand faces in a single image, which is not common in generic object detection. 
\begin{figure}[t]
    \begin{center}
        \includegraphics[width=\linewidth]{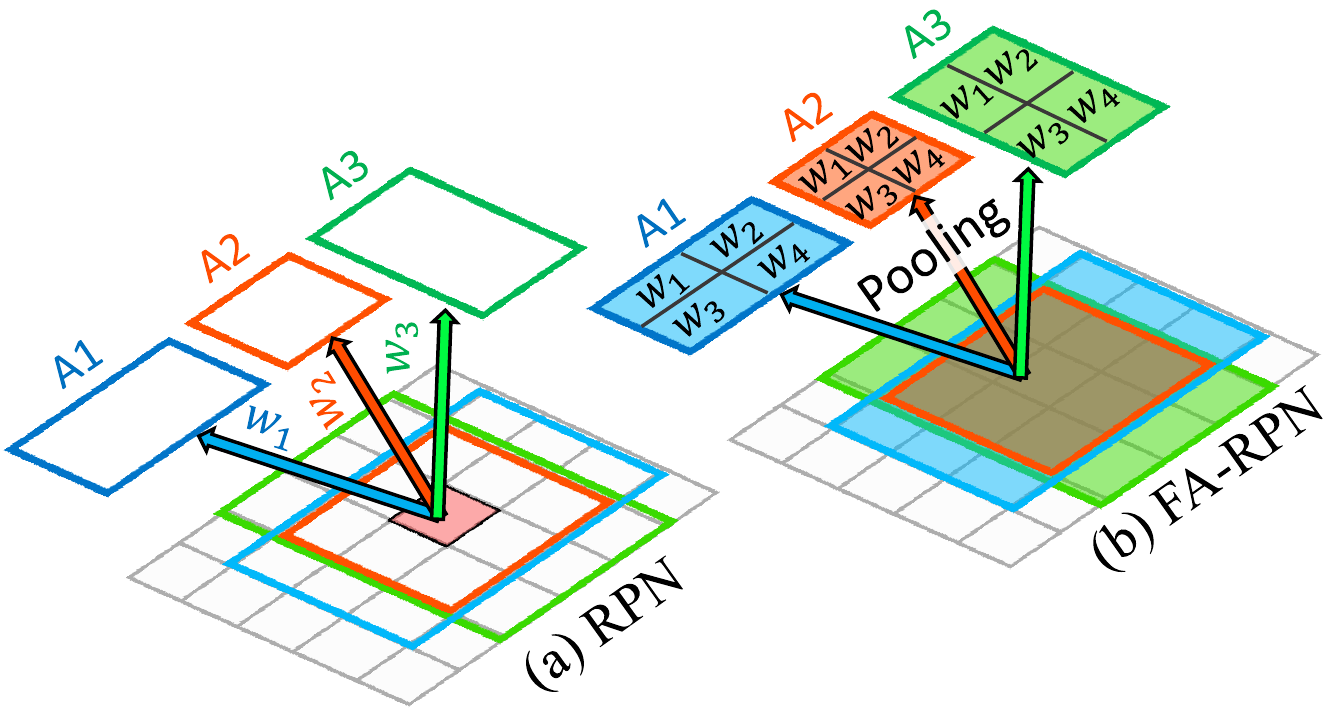}
    \end{center}
    \caption{Difference between RPN and FA-RPN in terms of weight configuration. For simplicity we show 2x2 pooling for FA-RPN.}
\label{fig:FARPN_weights}
\end{figure}

Detectors like Faster-RCNN \cite{ren2015faster} employ a region proposal network (RPN) which places anchor boxes of different sizes and aspect ratios uniformly on the image and classifies them for generating object-like regions. However, RPN only uses a single pixel in the convolutional feature map for evaluating the proposal hypotheses, independent of the size of the object. Therefore, the feature representation in RPN entirely relies on the contextual information encoded in the high-dimensional feature representation generated at the pixel. It does not pool features from the entire extent of an object while generating the feature representation, see Fig \ref{fig:FARPN_weights}. Thus, it can miss object regions or generate proposals which are not well localized. Further, it is not possible to iterate and refine the positions of the anchor-boxes as part of the proposal network. If objects of different scale/aspect-ratios are to be learned or if we want to place anchors at sub-pixel resolution, filters specific to each of these conditions need to be added during training. Generating proposals using a pooling based algorithm can alleviate such problems easily.

There are predominantly two pooling based methods for the final classification of RoIs in an image - Fast-RCNN \cite{girshick2015fast} and R-FCN \cite{dai2016r}. Fast-RCNN projects the region-proposals to the convolutional feature-map, and pools the features inside the region of interest (RoI) to a fixed size grid (typically 7$\times$7) and applies two fully connected layers which perform classification and regression. Due to computational constraints, this approach is practically infeasible for proposal generation as one would need to apply it to hundreds of thousands of regions - which is the number of region candidates which are typically evaluated by a region proposal algorithm.

To reduce the dependence on fully connected layers, R-FCN performs local convolutions (7$\times$7) inside an RoI for capturing the spatial-extent of each object. Since each of these local filters can be applied to the previous feature-map, we just need to pool the response from the appropriate region corresponding to each local filter. This makes it a good candidate for a pooling-based proposal approach as it is possible to apply it to a large number of RoIs efficiently. However, in high resolution images, proposal algorithms like RPN evaluate hundreds of thousands of anchors during inference. It is computationally infeasible to perform pooling on that many regions. Luckily, many anchors are not necessary (\eg large anchors which are very close to each other). In this paper, we show that careful anchor placement strategies can reduce the number of proposals significantly to the point where a pooling-based algorithm becomes feasible for proposal generation. This yields an efficient and effective objectness detector which does not suffer from the aforementioned problems present in RPN designs. 

\begin{figure*}[t]
    \begin{center}
        \includegraphics[width=\linewidth]{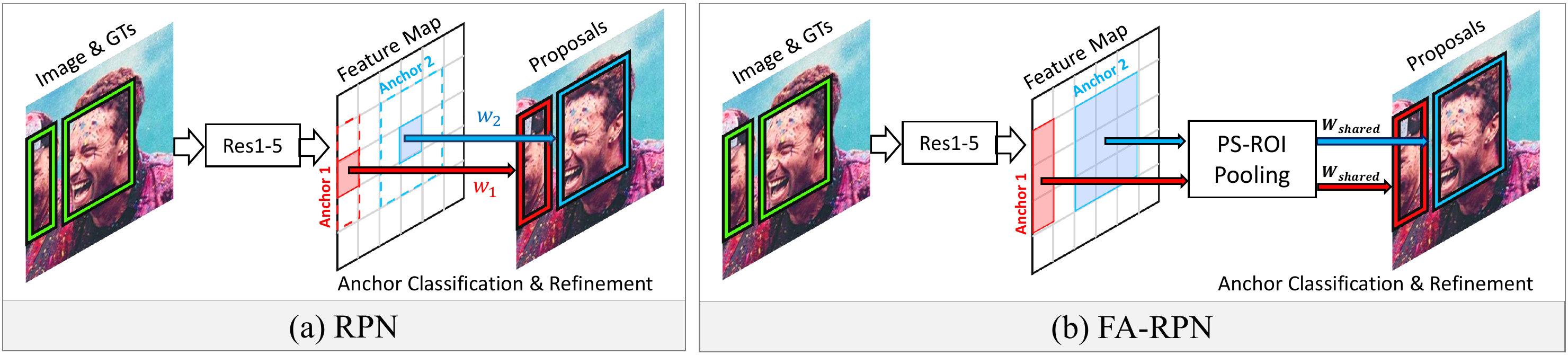}
    \end{center}
    \caption{We highlight the major differences between RPN (a) and FA-RPN (b) proposals. RPN performs classification on a single pixel in the high-dimensional feature-map and uses different weights for classifying anchor-boxes of different sizes/aspect ratios. FA-RPN proposals, on the other hand, pool features from multiple bins in the image and share weights across objects of different sizes and aspect ratios.}
\label{fig:diff}
\end{figure*}

A pooling-based proposal method based on R-FCN which relies on position sensitive filters is particularly well suited for face detection. While objects deform and positional correspondence between different parts is often lost - faces are rigid, structured and parts have positional semantic correspondence (\eg nose, eyes, lips). Moreover, it is possible to place anchor boxes of different size and aspect ratios without adding more filters. We can also place fractional anchor boxes and perform bilinear interpolation while pooling features for computing objectness. We can further improve localization performance of the proposal candidates by iteratively pooling again from the generated RoIs and all these design changes can be made during inference! Due to these reasons, we refer to our proposal network as {\em Floating Anchor Region Proposal Network} (FA-RPN). We highlight these advantages in Fig. \ref{fig:FARPN_weights} and Fig. \ref{fig:diff}. On the WIDER dataset \cite{yang2016wider} we show that FA-RPN proposals are better than RPN proposals. FA-RPN also obtains state-of-the-art results on WIDER and PascalFaces which demonstrates its effectiveness for face detection.

\section{Related Work}
Generating class agnostic region proposals has been investigated in computer vision for more than a decade. Initial methods include multi-scale combinatorial grouping \cite{arbelaez2014multiscale}, constrained parametric min-cuts \cite{uijlings2013selective}, selective search \cite{carreira2010constrained} \etc. These methods generate region proposals which obtain high recall for objects in a category agnostic fashion. They were also very successful in the pre-deep learning era and obtained state-of-the-art performance even with a bag-of-words model \cite{uijlings2013selective}. Using region proposals based on selective search \cite{uijlings2013selective}, R-CNN \cite{girshick2014rich} was the first deep learning based detector. Unsupervised region proposals were also used in later detectors like Fast-RCNN \cite{girshick2015fast} but since the Faster-RCNN detector \cite{ren2015faster} generated region proposals using a convolutional neural network, it has become the {\em de-facto} algorithm for generating region proposals.

To improve RPN, several modifications have been proposed. State-of-the-art detectors can also detect objects in a single step. Detectors like SSH \cite{najibi2017ssh}, SSD \cite{liu2016ssd}, RetinaNet \cite{lin2017focal}, MS-CNN \cite{cai2016unified} generate multi-scale feature maps to classify and regress anchors placed on these feature-maps. These single-shot detectors are closely related to the region proposal network as they have specific filters to detect objects of different sizes and aspect ratios but also combine feature-maps from multiple layers of the deep neural network. No further refinement is performed after the initial offsets generated by the network are applied. Another class of detectors are iterative, like G-CNN \cite{najibi2016g}, Cascade-RCNN \cite{cai2017cascade}, LocNet \cite{gidaris2016locnet}, FPN \cite{lin2017feature}, RFCN-3000 \cite{singh2017r}, Faster-RCNN \cite{ren2015faster}. These detectors refine a pre-defined set of anchor-boxes in multiple stages and have more layers to further improve classification and localization of regressed anchors. One should note that even in these networks, the first stage comprises of the region proposal network which eliminates the major chunk of background regions. FA-RPN is closer to this line of work but, in contrast, it supports iterative refinement of {\em region proposals} during inference.

We briefly review some recent work on face detection. With the availability of large scale datasets like WIDER \cite{yang2016wider} which contain many small faces in high resolution images, multiple new techniques for face detection have been proposed \cite{yang2015facial,zhang2017s,hu2017finding,zhu2017cms,li2016face,sun2017face,zhu2018seeing,ranjan2017hyperface,bai2018finding}. A lot of focus has been on scale, combining features of different layers \cite{hu2017finding,zhu2018seeing,najibi2017ssh,zhang2017s} and improving configurations of the region proposal network \cite{zhu2018seeing,zhang2017s}. For example, in finding tiny faces \cite{hu2017finding}, it is proposed to perform detection on an image pyramid and have different scale filters for objects of different sizes. SSH \cite{najibi2017ssh} and S3FD \cite{zhang2017s} efficiently utilize all the intermediate layers of the network. PyramidBox \cite{tang2018pyramidbox} replaces the context module in SSH by deeper and wider sub-networks to better capture the contextual information for face detection. Recently, even GANs \cite{goodfellow2014generative} have been used to improve the performance on tiny faces \cite{bai2018finding}.  

In face detection, the choice of anchors and their placement on the image is very important \cite{zhu2018seeing,zhang2017s}. For example, using extra strided anchors were shown to be beneficial \cite{zhu2018seeing}. Geometric constraints of the scene have also been used to prune region proposals \cite{amin2017geometric}. Some of these changes require re-training RPN again. In our framework, design decisions such as evaluating different anchor scales, changing the stride of anchors, and adding fractional anchors can simply be made during inference as we share filters for all object sizes and only pooling is performed for them. Moreover, a pooling based design also provides precise spatial features. 

\section{Background}
We provide a brief overview of the R-FCN detector in this section. This detector uses RPN to generate region proposals. It classifies the top 2000 ranked proposals using the R-FCN detector. Classification is performed over all the foreground classes and the background class. The key-component in R-FCN is local convolutions. It applies different filters in different sub-regions of an RoI for inferring the spatial extent of an object. These sub-regions may correspond to parts of an object. To accelerate this process over thousands of RoIs, convolution for each part in each object class is performed in the final layer. So, as an example, if there are 21 classes, the last feature-map would contain 21 $\times$ 49 channels. Then, given an RoI, {\em Position Sensitive RoIPooling} is performed on this feature-map to obtain the effect of local convolutions \cite{dai2016r}. We refer the reader to the R-FCN \cite{dai2016r} paper for further details on PSRoIPooling. Finally, the response is average pooled and used as the classification score of the object. In Deformable-RFCN \cite{dai2017deformable}, the regions for each bin where pooling is performed are also adjusted based on the input feature-map, which is referred to as deformable PSRoIPooling.

\section{FA-RPN - Floating Anchor Region Proposal Network}
\begin{figure*}
    \begin{center}
        \includegraphics[width=0.85\linewidth]{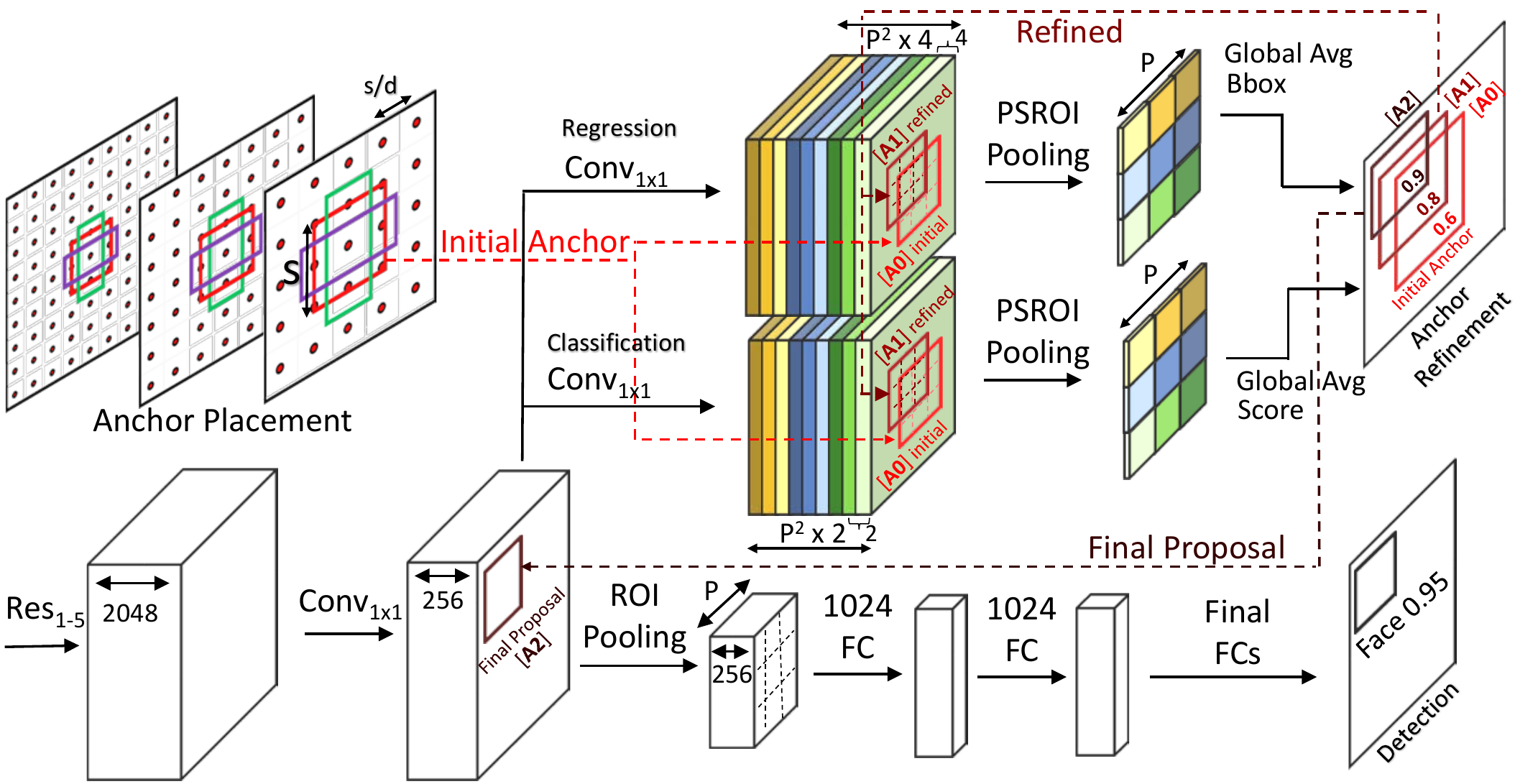}
    \end{center}
    \caption{FA-RPN framework. FA-RPN uses multi-scale training. At each training iteration, an image scale is randomly selected and \emph{suitable} anchor scales are placed over the image. This set of initial anchors are used to pool objectness scores and localization information from position sensitive filters (for simplicity only the localization branch is depicted in the figure). For improving localization, the top scoring anchors is further refined with subsequent poolings. Finally, a Faster-RCNN head is used to perform the final classification and regression.}
\label{fig:framew}
\end{figure*}

In this section, we discuss training of FA-RPN, which performs iterative classification and regression of anchors placed on an image for generating accurate region proposals. An overview of our approach is shown in Fig. \ref{fig:framew}.

\subsection{Anchor Placement}
In this architecture, classification is not performed using a single high-dimensional feature vector but by pooling features inside the RoI. Hence, there are no restrictions on how RoIs can be placed during training and inference. As long as the convolutional filters can learn objectness, we can apply the model on RoIs of different sizes and aspect ratios, even if the network was not trained explicitly for those particular scales and aspect-ratios.

FA-RPN places anchors of different scales and aspect ratios on a grid, as generated in the region proposal network, and clips the anchors which extend beyond the image. While placing anchors, we vary the spatial stride as we increase the anchor size. Since nearby anchors at larger scales have a very high overlap, including them is not necessary. We change the stride of anchor-boxes to $max(c, s/d)$, where $s$ is square-root of the area of an anchor-box, $c$ is a constant and $d$ is the scaling factor, shown in Fig \ref{fig:framew}. In practice, we set $c$ to 16 and $d$ to 5. This ensures that not too many overlapping anchor-boxes are placed on the image, while ensuring significant overlap between adjacent anchors to cover all objects. {\em Naive placement of anchor boxes of 3 aspect ratios and 5 scales with stride equaling 16 pixels in a 800 $\times$ 1280 image leads to a 2-3 $\times$ slow-down when performing inference}. With the proposed placement method, we reduce the number of RoIs per image from 400,000 to 100,000 for a 1280 $\times$ 1280 image for the above mentioned anchor configuration. When we increase the image size, computation for convolution also increases proportionally, so as long as the time required for pooling is not significant compared to convolution, we will not observe a noticeable difference in performance.

There is no restriction that the stride of anchors should be the same as the stride of the convolutional feature-map. We can even place RoIs between two pixels in the convolutional feature-map without making any architectural change to the network. This allows us to augment the ground-truth bounding boxes as positive RoIs during training. This is unlike RPN, where the maximum overlapping anchor is assigned as positive when no anchor matches the overlap threshold criterion. We show qualitative examples of anchor placement for different scales and aspect ratios in FA-RPN in Fig. \ref{fig:framew}. 

\subsection{Sampling}
Since there are hundreds of thousands of anchors which can be placed on an image, we sample anchors during training. We observe that using focal loss \cite{lin2017focal} reduced recall for RPN (hyper-parameter tuning could be a reason), so we did not use it for FA-RPN. We use the commonly used technique of sampling RoIs for handing class imbalance. In FA-RPN, an anchor-box is marked as positive if its overlap with a ground truth box is greater than 0.5. An anchor is marked as negative if its overlap is less than 0.4. A maximum of 128 positive and negative anchors are sampled in a batch. Since the probability of a random anchor being an easy sample is high, we also sample 32 anchor-boxes which have an overlap of at-least 0.1 with the ground-truth boxes as hard negatives. Just for training FA-RPN proposals, all other RoIs can be ignored. However, for training an end-to-end detector, we also need to score other RoIs in the image. When training an end-to-end detector, we select a maximum of 50,000 RoIs in an image (prioritizing those which have at-least 0.1 overlap with ground-truth boxes first).


\subsection{Iterative Refinement}
The initial set of placed anchors are expected to cover the ground-truth objects present in the image. However, these anchors may not always have an overlap greater than 0.5 with all objects and hence would be given low scores by the classifier. This problem is amplified for small object instances as mentioned in several methods \cite{zhang2017s,hu2017finding}. In this case, no anchor-boxes may have a high score for some ground-truth boxes. Therefore, the ground-truth boxes may not be covered in the top 500 to 1000 proposals generated in the image. In FA-RPN, rather than selecting the top 1000 proposals, we generate 20000 proposals during inference and then perform pooling again on these 20000 proposals from the same feature-map (we can also have another convolutional layer which refines the first stage region proposals). The hypothesis is that after refinement, the anchors would be better localized and hence the scores which we obtain after pooling features inside an RoI would be more reliable. Therefore, after refinement, the ordering of the top 1000 proposals would be different because scores are pooled from refined anchor-boxes rather than the anchor-boxes which were placed uniformly on a grid. Since we only need to perform pooling for this operation, it is efficient and can be easily implemented when the number of RoIs is close to 100,000. Note that our method is entirely pooling based and does not have any fully connected layers like cascade-RCNN \cite{cai2017cascade} or G-CNN \cite{najibi2016g}. Therefore, it is much more efficient for iterative refinement.

\subsection{Complexity and Speed}
FA-RPN is very efficient. On 800 $\times$ 1280 size images, it takes 50 milliseconds to perform forward propagation for our network on a P6000 GPU. We also discuss how much time it takes to use R-FCN for end-to-end detection. For general object detection, when the number of classes is increased, to say 100, the contribution from the pooling layer also increases. This is because the complexity for pooling is linear in the number of classes. So, if we increase the number of classes to 100, this operation would become 100 times slower and at that stage, pooling will account for a significant portion of the time in forward-propagation. For instance, without our anchor placement strategy, it takes 100 seconds to perform inference for 100 classes in a single image on a V100 GPU. However, as for face detection, we only need to perform pooling for 2 classes and use a different anchor placement scheme, we do not face this problem and objectness can be efficiently computed even with tens of thousands of anchor boxes.

\subsection{Scale Normalized Training}
The positional correspondence of R-FCN is lost when RoI bins become too small. The idea of local convolution or having filters specific to different parts of an object is relevant when each bin corresponds to a unique region in the convolutional feature-map. The position-sensitive filters implicitly assume that features in the previous layer have a resolution which is similar to that after PSRoIPooling. Otherwise, if the RoI is too small, then all the position sensitive filters will pool from more or less the same position, nullifying the hypothesis that these filters are position sensitive. Therefore, we perform scale normalized training \cite{singh2017analysis}, which performs selective gradient propagation for RoIs which are close to a resolution of 224 $\times$ 224 and excludes those RoIs which can be observed at a better resolution during training. In this setting, the position-sensitive nature of filters is preserved to some extent, which helps in improving the performance of FA-RPN. 

\section{Datasets}
We perform experiments on three benchmark datasets, WIDER \cite{yang2016wider}, AFW \cite{zhu2012face}, and Pascal Faces \cite{yan2014face}. The WIDER dataset contains 32,203 images with 393,703 annotated faces, 158,989 of which are in the train set, 39,496 in the validation set, and the rest are in the test set. The validation and test set are divided into ``easy", ``medium", and ``hard" subsets cumulatively (i.e. the ``hard" set contains all faces and ``medium" contains ``easy" and ``medium"). This is the most challenging public face dataset mainly due to the significant variation in the scale of faces and occlusion. We train all models on the train set of the WIDER dataset and evaluate on the validation set. We mention in our experiments when initialization of our pre-trained model is from ImageNet or COCO. Ablation studies are also performed on the the validation set (i.e. ``hard" subset which contains the whole dataset). Pascal Faces and AFW have 1335 and 473 faces respectively. We use Pascal Faces and AFW only as test sets for evaluating the generalization of our trained models. When performing experiments on these datasets, we apply the model trained on the WIDER train set out of the box. 

\section{Experiments}
\label{sec:exp}
We train a ResNet-50 \cite{he2016deep} based Faster-RCNN detector with deformable convolutions \cite{dai2017deformable} and SNIP \cite{singh2017analysis}. FA-RPN proposals are generated on the concatenated \textit{conv4} and \textit{conv5} features. On WIDER we train on the following image resolutions $(1800,2800)$, $(1024, 1440)$ and $(512, 800)$. The SNIP ranges we use for WIDER are as follows, [0, 200) for (1800, 2800), [32, 300) for (1024, 1440) and [80, $\infty$) for (512, 800) as the size of the shorter side of the image is around 1024. We train for 8 epochs with a stepdown at 5.33 epochs. In all experiments we use a learning rate and weight decay of 0.0005 and train on 8 GPUs. We use the same learning rate and training schedule even when training on 4 GPUs. In all our experiments, we use online hard example mining \textit{(OHEM)} \cite{shrivastava2016training} to train the 2 fully connected layers in our detector. For the detector, we perform hard example mining on 900 proposals with a batch size of 256. RoIs greater than 0.5 overlap with ground-truth bounding boxes are marked as positive and anything less than that is labelled as negative. No hard-example mining is performed for training the Faster-RCNN head. We use Soft-NMS \cite{bodla2017soft} with $\sigma = 0.35$ when performing inference. Since Pascal Faces and AFW contain low resolution images and also do not contain faces as small as the WIDER dataset, we do not perform inference on the $1800\times2800$ resolution. All other parameters remain the same as the experiments on the WIDER dataset. 


On the WIDER dataset, we remove anchors for different aspect ratios (i.e. we only have one anchor per scale with an aspect ratio of 1) and add a $16\times16$ size anchor for improving the recall for small faces. Note that extreme size anchors are removed during training with SNIP using the same rules which are used for training Faster-RCNN. With these settings, we outperform state-of-the-art results on the WIDER dataset demonstrating the effectiveness of FA-RPN. However, the objective of this paper is not to show that FA-RPN is necessary to obtain state-of-the-art performance. FA-RPN is an elegant and efficient alternative to RPN and can be combined with multi-stage face detection methods to improve performance.

\begin{figure}
    \centering
    \includegraphics[width=0.85\linewidth]{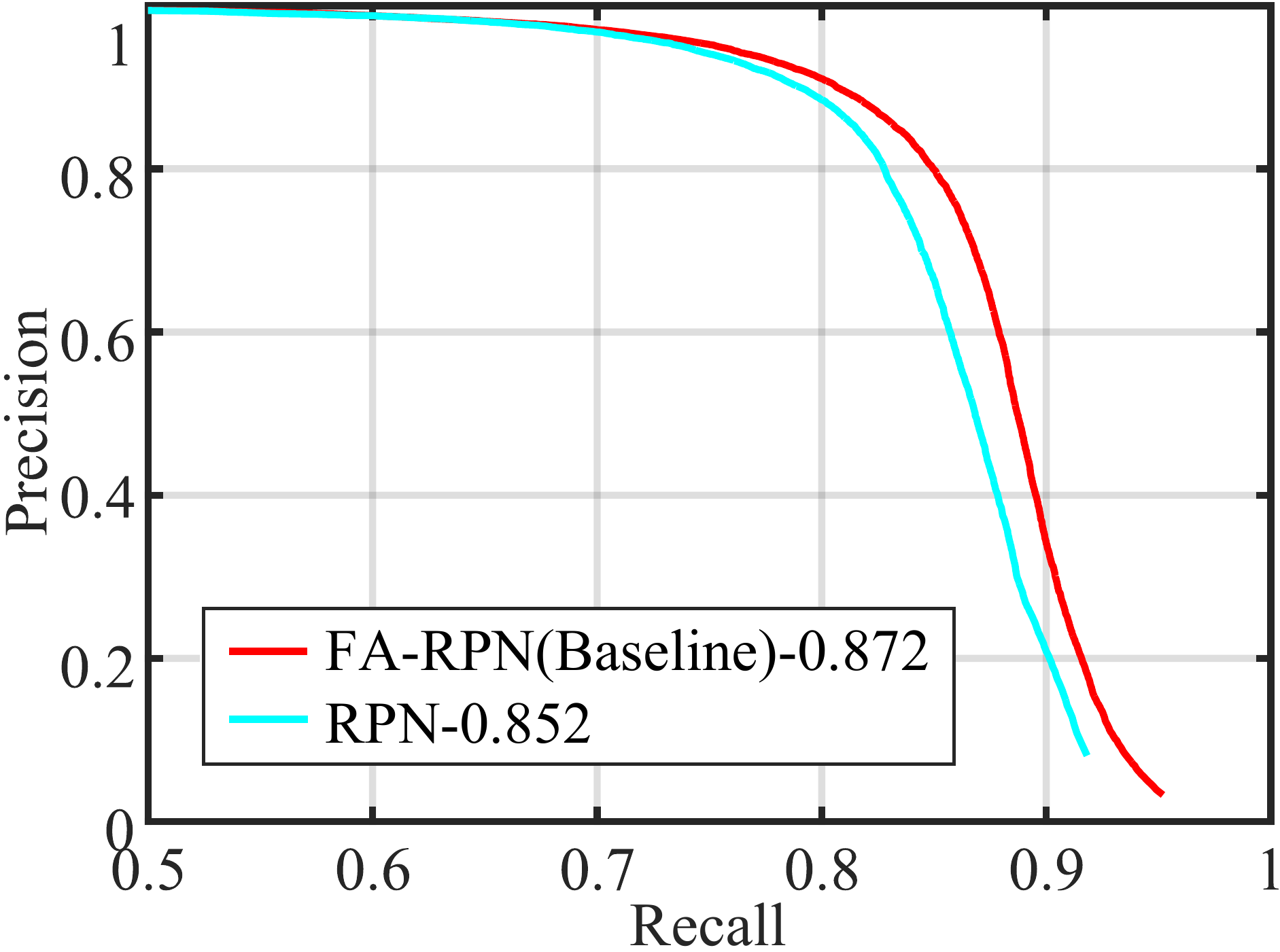}
    \caption{Comparison with RPN on the WIDER dataset.}
    \label{fig:rpn_comparison_wider}
\end{figure}


  

\begin{table}
	\begin{center}
		\begin{tabular}{l|c}
			\hline
			Method & AP \\
			\hline
			\small{Baseline} & 87.2 \\
			\small{Baseline + SNIP} & 88.1\\
			\small{Baseline + SNIP + COCO pre-training} & 89.1 \\
			\small{Baseline + SNIP + COCO pre-training + Iteration} & 89.4\\
			\hline
		\end{tabular}
	\end{center}
	\caption{Ablation analysis with different core-components of our face detector on the hard-set of the WIDER dataset (hard-set contains all images in the dataset).}
	\label{tb:det}
\end{table}

\subsection{Effect of Multiple Iterations in FA-RPN}
\begin{figure}
    \centering
    \begin{subfigure}[t]{0.505\linewidth}
    \includegraphics[width=\linewidth]{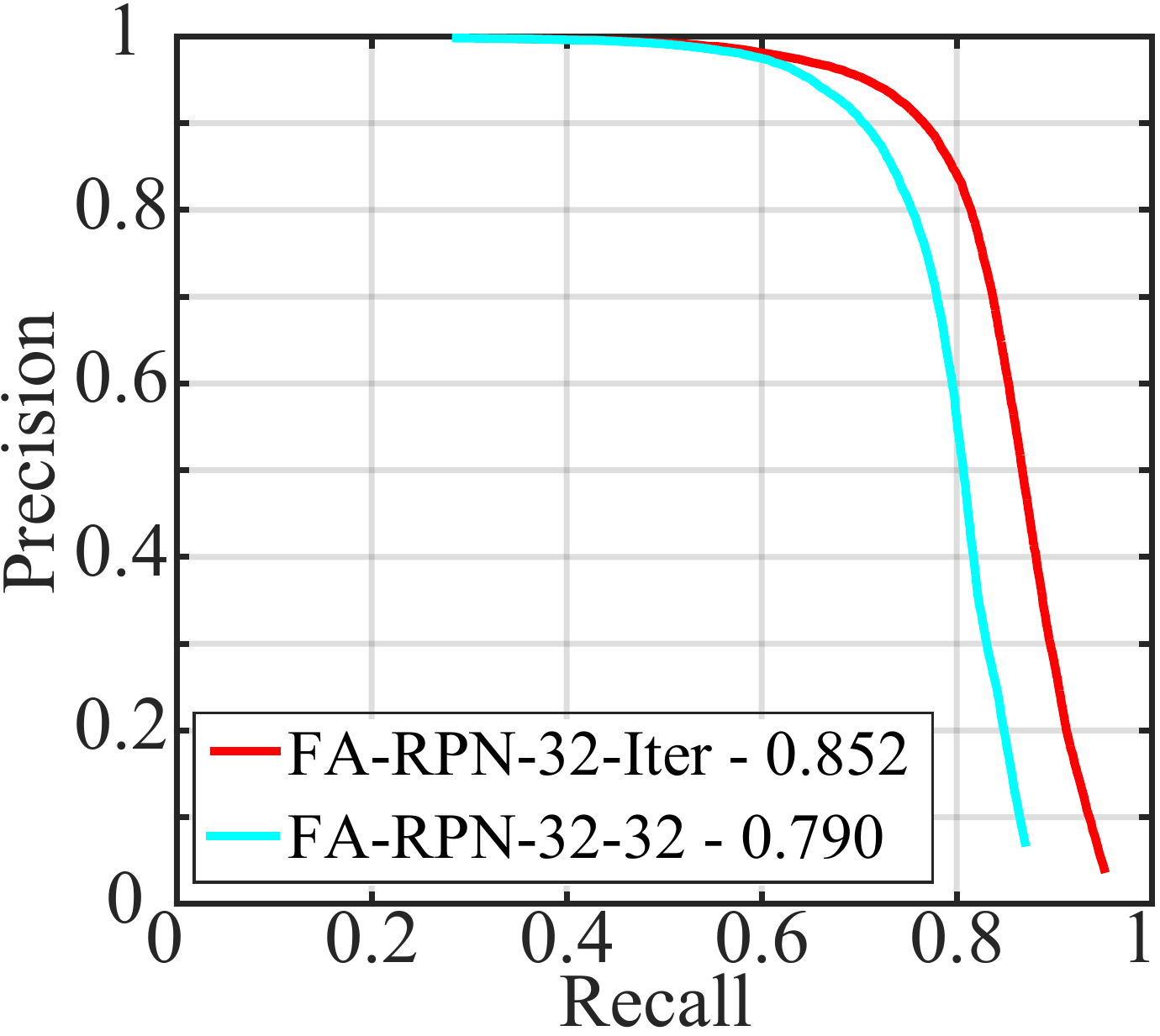}
    \caption{}
    \end{subfigure}
    \begin{subfigure}[t]{0.485\linewidth}
    \includegraphics[width=\linewidth]{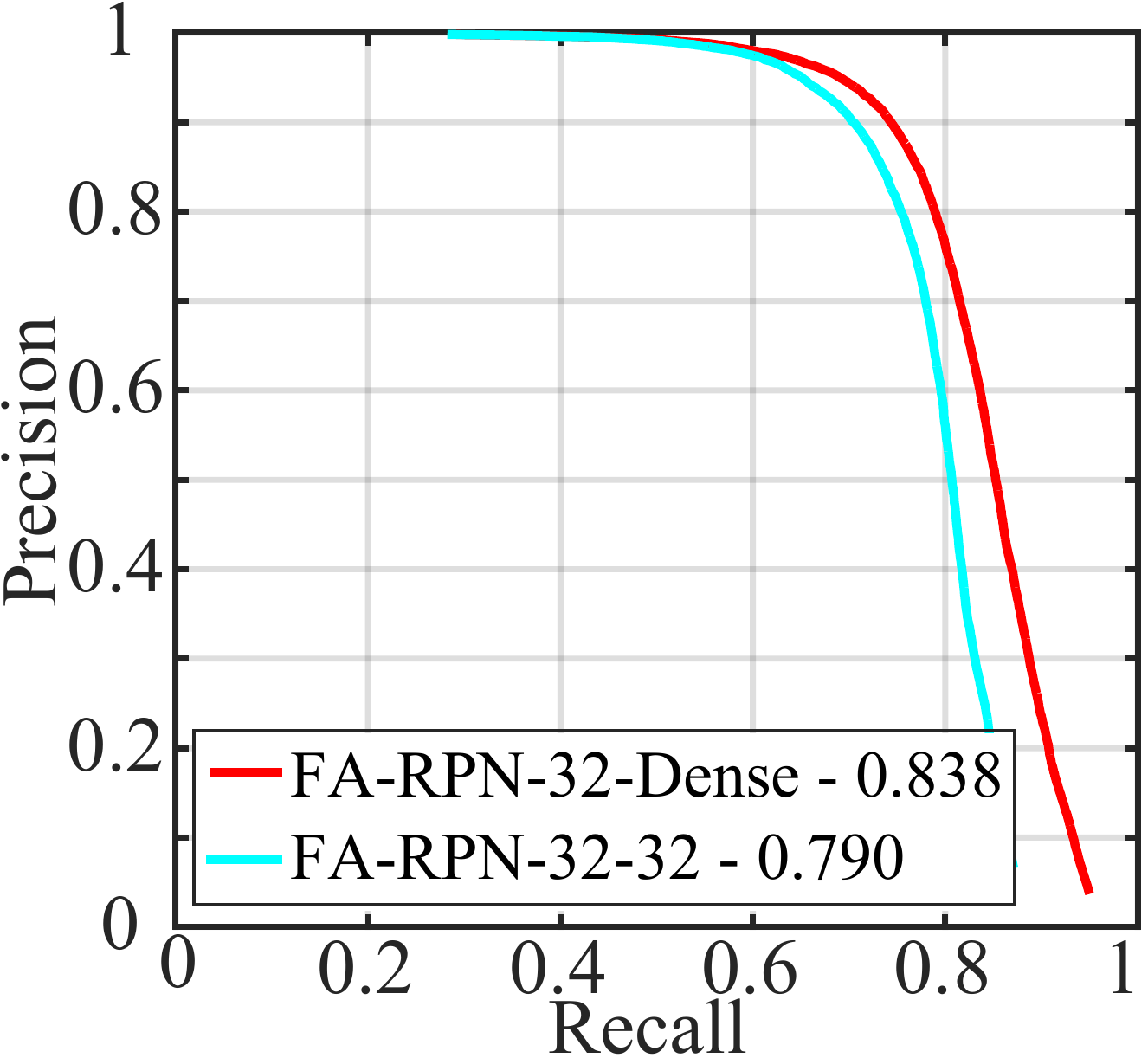}
    \caption{}
    \end{subfigure}
    \caption{Ablation analysis: improving precision at inference time. \textit{FA-RPN-32-32} represents a model which is trained by increasing the stride between anchors to 32 and uses the same stride at inference time. (a) \textit{FA-RPN-32-Iter} is the same model when an additional anchor refinement step is performed at {\em inference}. (b) \textit{FA-RPN-32-Dense} on the other hand, improves precision by reducing the anchor stride at inference time to our original FA-RPN stride.}
\label{fig:expab}
\end{figure}

We evaluate FA-RPN on WIDER when we perform multiple iterations during inference. Since FA-RPN operates on RoIs rather than classifying single-pixel feature-maps like RPN, we can further refine the RoIs which are generated after applying the regression offsets. As the initial set of anchor boxes are coarse, the RoIs generated after the first step are not very well localized. Performing another level of pooling on the generated RoIs helps to improve recall for our proposals. As can be seen in Table \ref{tb:det} and the left-hand side plot in Fig. \ref{fig:expab}, this refinement step helps to improve the precision and recall. We also generate anchors with different strides - 16 and 32 pixels - and show how the final detection performance improves as we refine proposals. 


\subsection{Evaluating different Anchors and Strides during Inference}
In this section, we show the flexibility of FA-RPN for generating region proposals. We train our network with a stride of $32$ pixels and during inference, we generate anchors at a stride of $16$ pixels on the WIDER dataset. The result is shown in the right-hand side plot in Fig. \ref{fig:expab}. We notice that the dense anchors improve performance by 3.8\%. On the left side of the plot we show the effect of iterative refinement of FA-RPN proposals. This further provides a boost of 1.4\% on top of the denser anchors. This shows that our network is robust to changes in anchor configuration, and can detect faces even on anchor sizes which were not provided during training. To achieve this with RPN, one would need to re-train it again, while in FA-RPN it is a simple inference time hyper-parameter which can be tuned on a validation set even after the training phase. 

\subsection{Effect of Scale and COCO pre-training on Face Detection}
Variation of scale is among the main challenges in detection datasets. Datasets like WIDER consist of many small faces which can be hard to detect for a CNN at the original image scale. Therefore, upsampling images is crucial to obtaining good performance. However, as shown in \cite{singh2017analysis}, when we upsample images, large objects become hard to classify and when we downsample images to detect large objects, small objects become harder to classify. Therefore, standard multi-scale training is not effective when training on extreme resolutions. In Table \ref{tb:det} we show the effect of performing SNIP based multi-scale training in our FA-RPN based Faster-RCNN detector. When performing inference on the same resolutions, we observe an  improvement in detection performance on the WIDER dataset by 1\%. Note that this improvement is on top of multi-scale inference. We also initialized our ResNet-50 model which was pre-trained on the COCO detection dataset. We show that even pre-training on object detection helps in improving the performance of face detectors by a significant amount, Table \ref{tb:det}.

\begin{figure*}
    \centering
    \begin{subfigure}[t]{0.32\linewidth}
    \includegraphics[width=\linewidth]{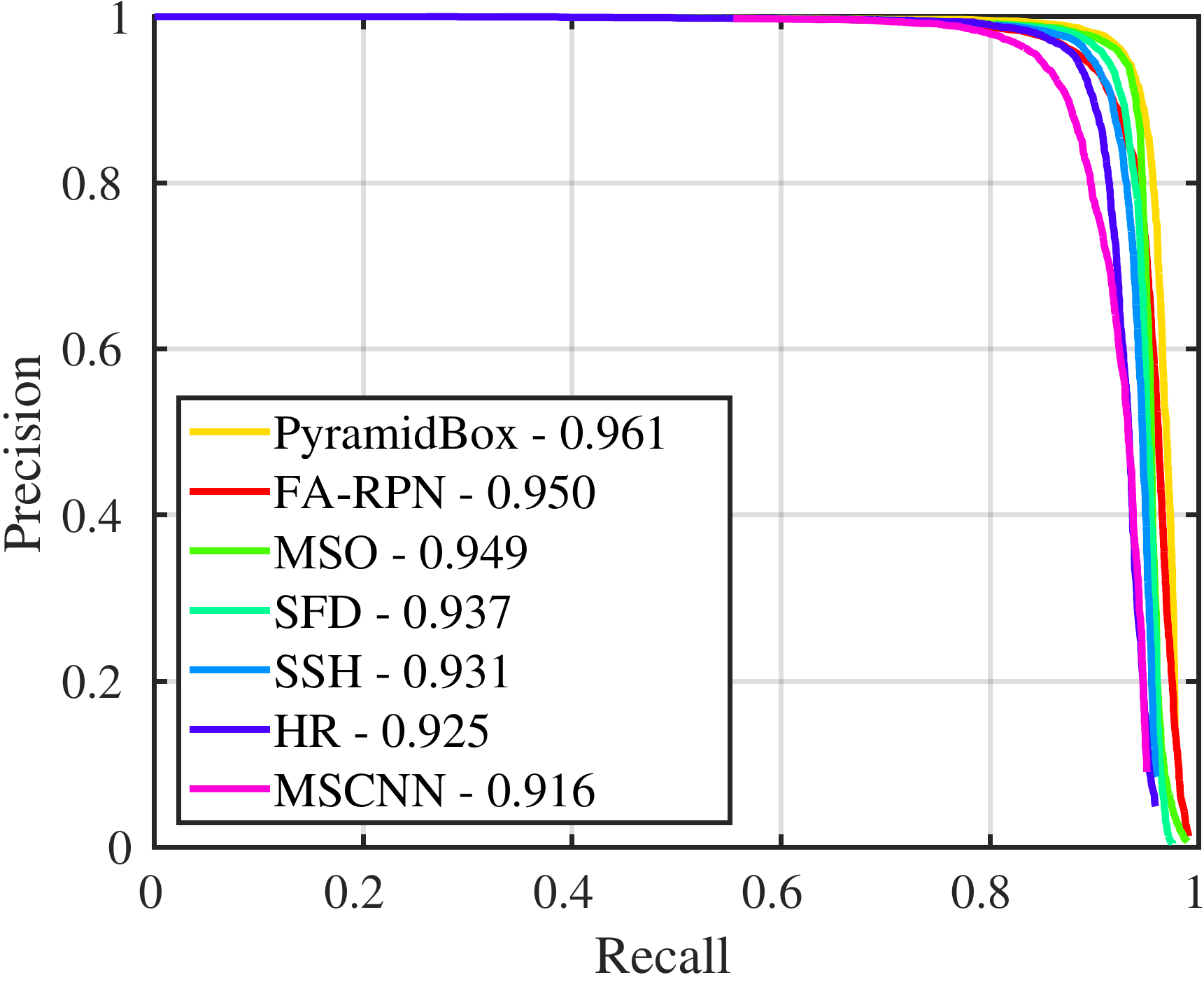}
    \caption{Easy}
    \end{subfigure}
    \begin{subfigure}[t]{0.32\linewidth}
    \includegraphics[width=\linewidth]{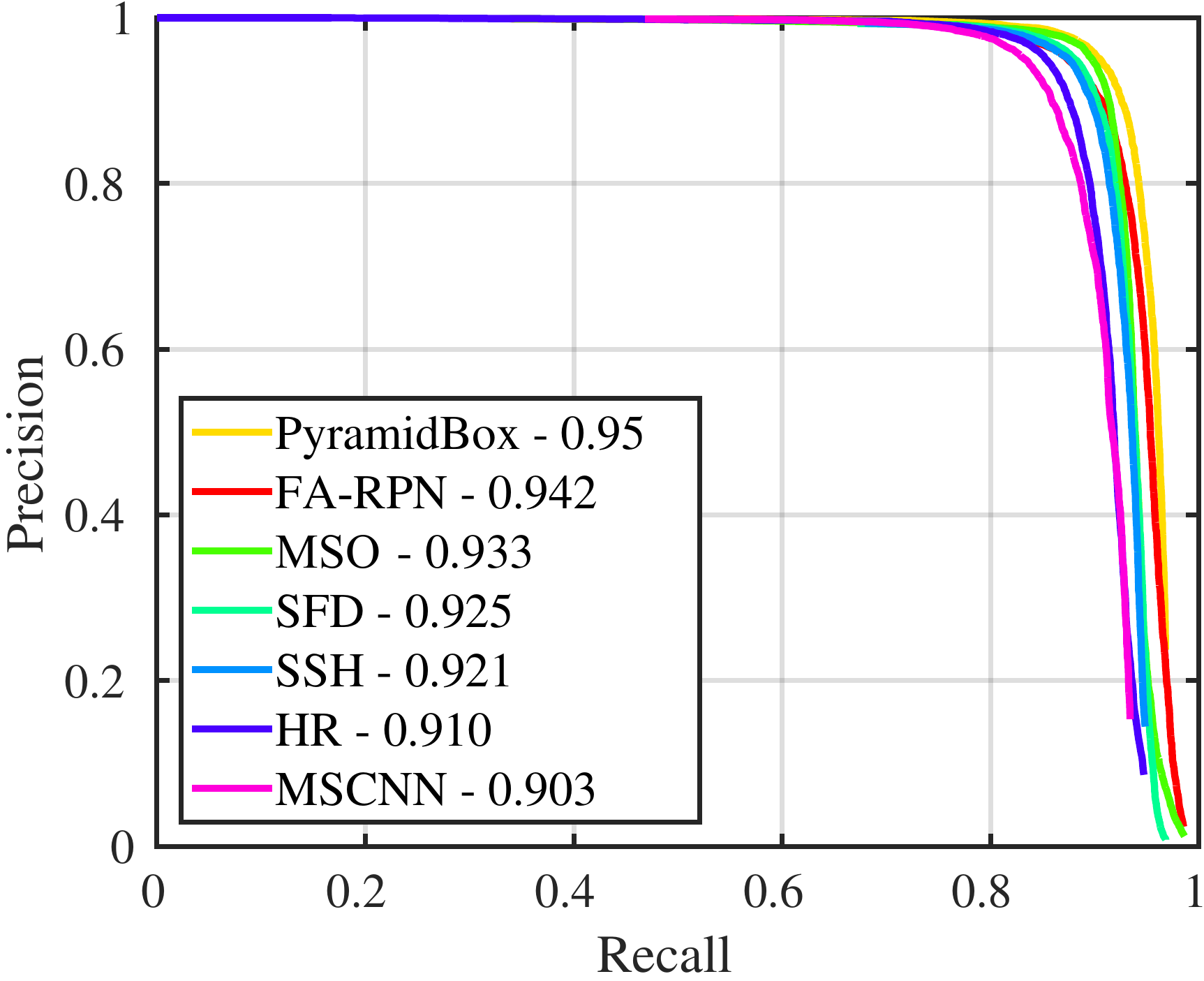}
    \caption{Medium}
    \end{subfigure}
      \begin{subfigure}[t]{0.32\linewidth}
    \includegraphics[width=\linewidth]{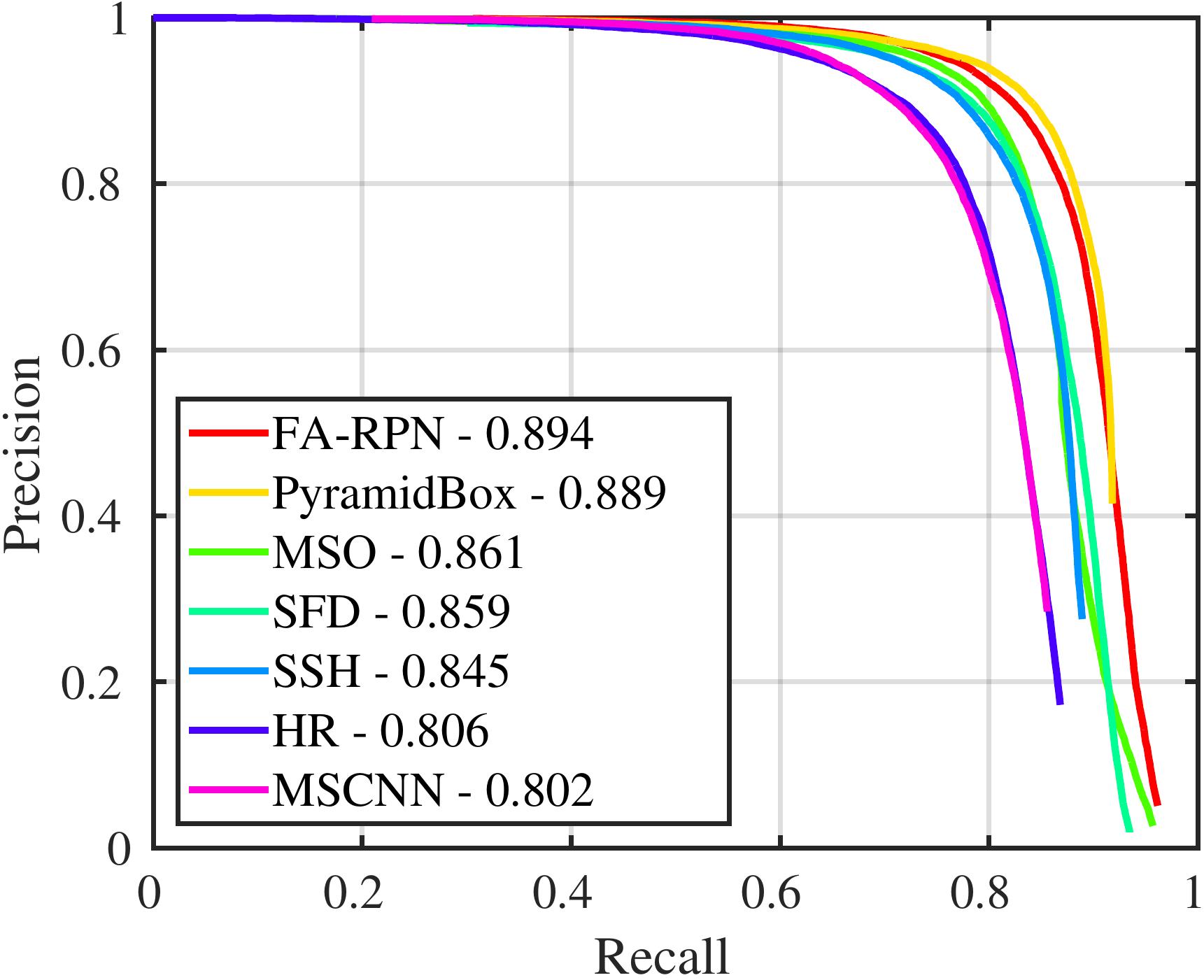}
    \caption{Hard (whole dataset)}
    \end{subfigure}
   
    \caption{We compare with recently published methods on the WIDER dataset. The plots are for ``easy", ``medium" and ``hard" respectively from left to right. As can be seen, FA-RPN outperforms published baselines on this dataset. Note that, ``hard'' set contains the whole dataset while ``easy'' and ``medium'' are subsets.}
    \label{fig:exp}
\end{figure*}



\begin{figure}
    \centering
    \begin{subfigure}[t]{0.49\linewidth}
    \includegraphics[width=\linewidth]{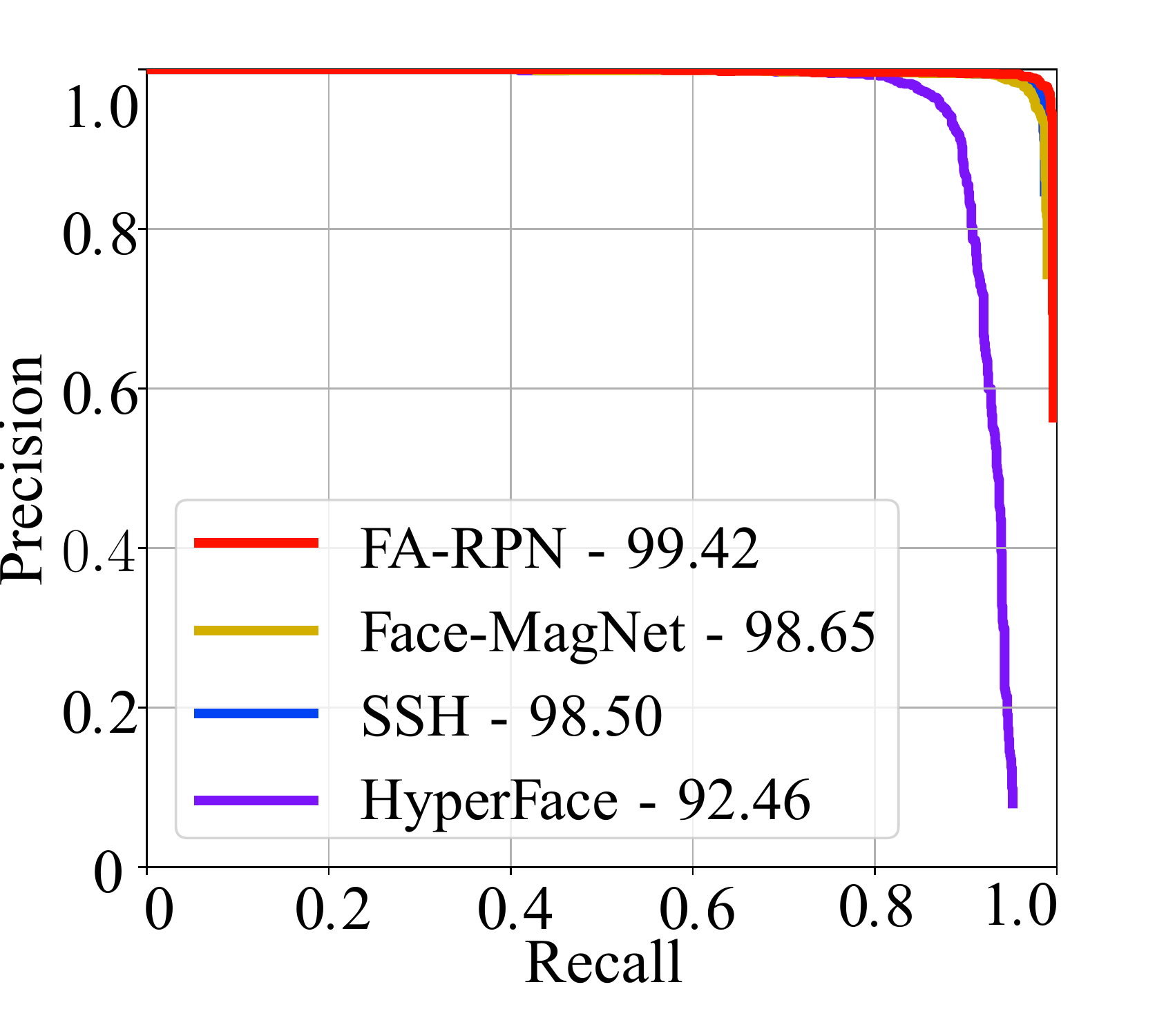}
    \caption{}
    \end{subfigure}
    \begin{subfigure}[t]{0.49\linewidth}
    \includegraphics[width=\linewidth]{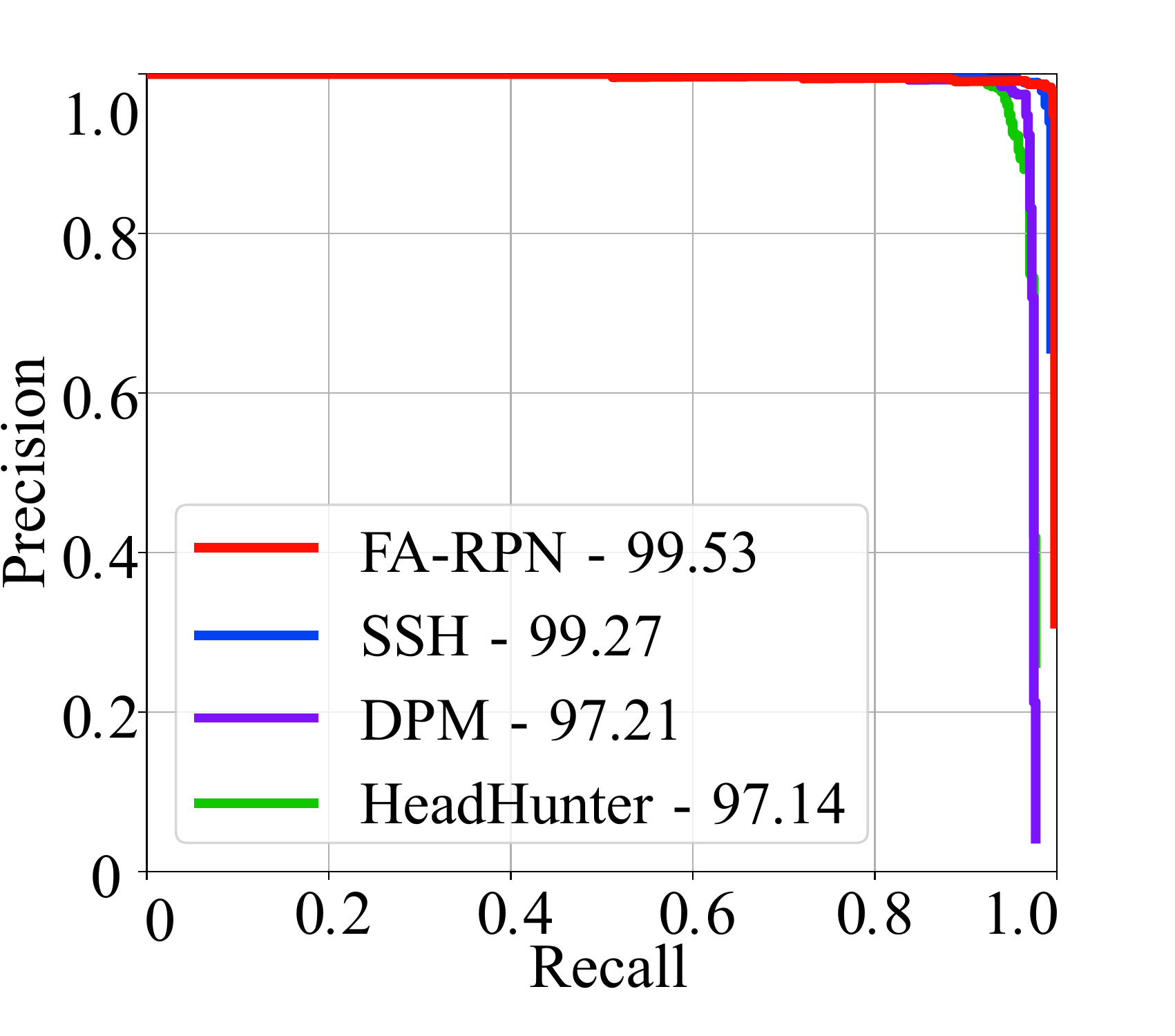}
    \caption{}
    \end{subfigure}
   
    \caption{Comparison with other methods on (a) Pascal Faces, and (b) AFW datasets.}
     \label{fig:pascal_faces}
\end{figure}
\subsection{Comparison on the WIDER dataset}
We compare our method with MSCNN \cite{cai2016unified}, HR \cite{hu2017finding}, SSH \cite{najibi2017ssh}, S3FD \cite{zhang2017s}, MSO \cite{zhu2018seeing}, and PyramidBox \cite{tang2018pyramidbox} which are the published state-of-the-art methods on the WIDER dataset. Our simple detector outperforms all existing methods on this dataset. On the ``hard'' set, which includes {\em all} the annotations in the WIDER dataset, our performance (average precision) is $89.4$\%, which is the best among all methods. We also perform well in the easy and medium sets. The precision recall plots for each of these cases are shown in Fig. \ref{fig:exp}.
Note that we did not use feature-pyramids or lower layer features from conv2 and conv3 \cite{najibi2017ssh,zhang2017s,hu2017finding} , enhancing predictions with context \cite{hu2017finding}  or with deeper networks like ResNext-152 \cite{xie2017aggregated}/ Xception \cite{chollet2016xception} for obtaining these results. This result demonstrates that FA-RPN is competitive with existing proposal techniques as it can lead to a state-of-the-art detector. We also do not use recently proposed techniques like stochastic face lifting \cite{zhu2018seeing}, having different filters for different size objects \cite{hu2017finding} or maxout background loss \cite{zhang2017s}. Our performance can be further improved if the above mentioned architectural changes are made to our network or better training methods which also fine-tune batch-normalization statistics are used \cite{peng2017megdet,singh2018sniper}. 



\begin{figure*}[b!]
    \begin{center}
        \includegraphics[width=1\linewidth]{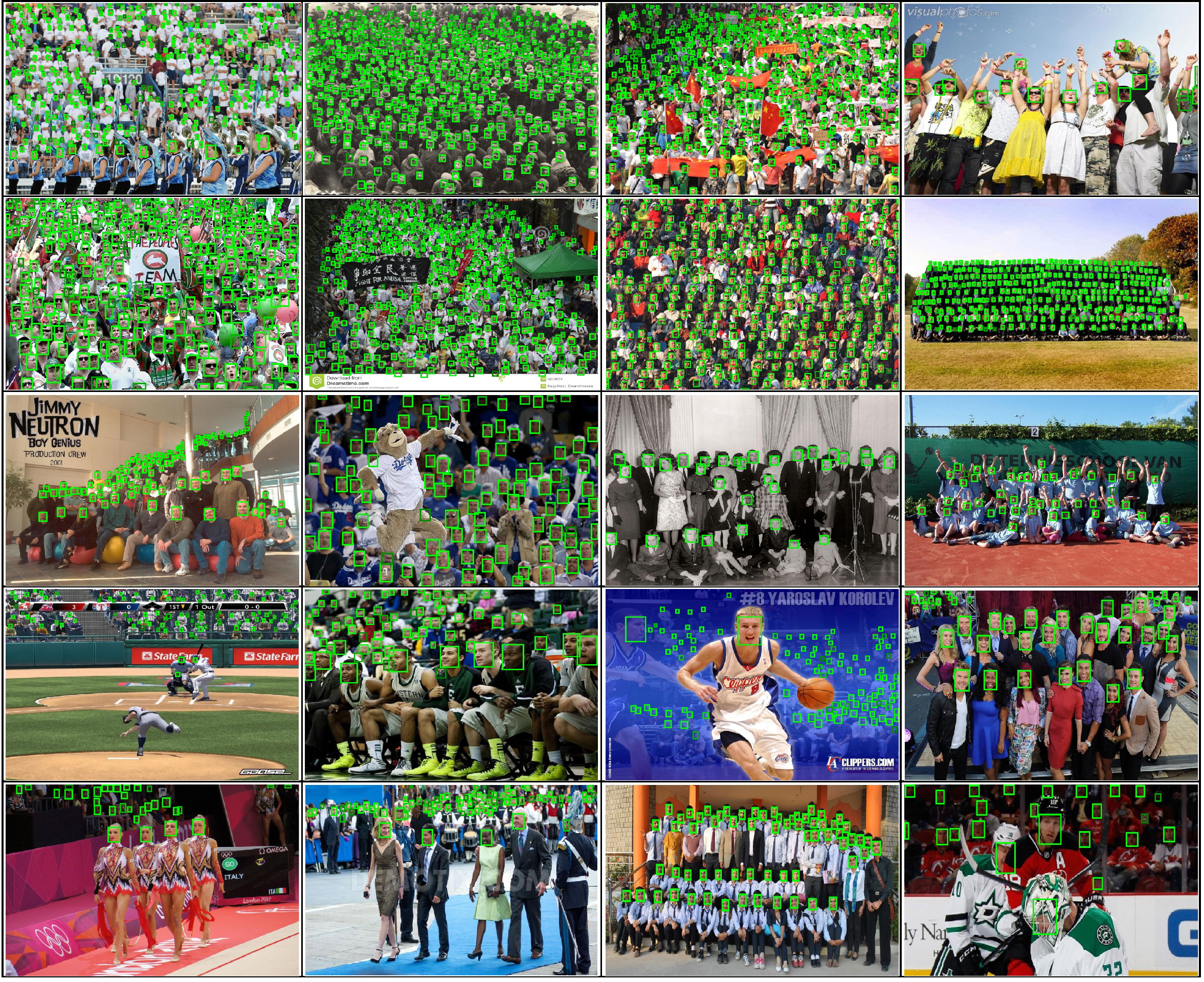}
    \end{center}
    \caption{Qualitative results on the validation set of the WIDER dataset. Green rectangles show the detection and brightness encodes the detection confidence.}
\label{fig:qual}
\end{figure*}
\subsection{Comparison on the PascalFaces and AFW datasets}
To show the generalization of our trained detector, we also apply it out-of-the-box to the Pascal Faces \cite{yan2014face} and AFW \cite{zhu2012face} datasets without fine-tuning. The performance of FA-RPN is compared with SSH \cite{najibi2017ssh}, Face-Magnet \cite{samangouei2018face}, HyperFace \cite{ranjan2017hyperface}, HeadHunter \cite{mathias2014face}, and DPM  \cite{ranjan2015deep} detectors which reported results on these datasets. The results are shown in Fig. \ref{fig:pascal_faces}. Compared to WIDER, the resolution of PASCAL images is lower and they do not contain many small images, so it is sufficient to apply FA-RPN to the two lower resolutions in the pyramid. This also leads to faster inference. As can be seen, FA-RPN out-of-the-box generalizes well to these datasets. FA-RPN achives state-of-the-art result on the PascalFaces and reduces the error rate to 0.68\% on this dataset.


\subsection{Efficiency}
Our FA-RPN based detector is efficient and takes less than $0.05$ seconds to perform inference on an image of size $800 \times 1280$. With advances in GPUs over the last few years, performing inference even at very high resolutions $(1800 \times 2800)$ is efficient and takes less than 0.4 seconds on a 1080Ti GPU. With improved GPU architectures like the 2080Ti and with the use of lower precision like 16 or 8 bits, the speed can be further improved by two to four times (depending on the precision used in inference) at the same cost. Multi-scale inference can be further accelerated with AutoFocus \cite{najibi2018autofocus}. 

\subsection{Qualitative Results}
Figure \ref{fig:qual} shows qualitative results on the WIDER validation subset. We picked 20 diverse images to highlight the results generated by FA-RPN. Detections are shown by green rectangles and the brightness encodes the confidence. As can be seen, our face detector works very well in crowded scenes and can find hundreds of small faces in a wide variety of images. This shows that FA-RPN have a very high recall and can detect faces accurately. It generalizes well in both indoor and outdoor scenes and under different lighting conditions. Our performance across a wide range of scales is also good without using diverse features from different layers of the network. It is also robust to changes in pose, occlusion, blur and even works on old photographs!

\section{Conclusion}
We introduced FA-RPN, a novel method for generating pooling based proposals for face detection. We proposed techniques for anchor placement and label assignment which were essential in the design of such pooling based proposal algorithm. FA-RPN has several benefits like efficient iterative refinement, flexibility in selecting scale and anchor stride during inference, sub-pixel anchor placement \etc. Using FA-RPN, we obtained state-of-the-art results on the challenging WIDER dataset, showing the effectiveness of FA-RPN for this task. FA-RPN also achieved state-of-the-art results out-of-the-box on datasets like PascalFaces showing its generalizability.

\clearpage
\small{\textbf{Acknowledgement} This research is based upon work supported by the Office of the Director of National Intelligence (ODNI), Intelligence Advanced Research Projects Activity (IARPA), via IARPA R\&D Contract No. 2014-14071600012. The views and conclusions contained herein are those of the authors and should not be interpreted as necessarily representing the official policies or endorsements, either expressed or implied, of the ODNI, IARPA, or the U.S. Government. The U.S. Government is authorized to reproduce and distribute reprints for Governmental purposes notwithstanding any copyright annotation thereon.}

{\small
\bibliographystyle{ieee}
\bibliography{egbib}
}
\end{document}